# Text-to-Battery Recipe: A language modeling-based protocol for automatic battery recipe extraction and retrieval

*Daeun Lee[1,2,+], Jaewoong Choi[1,+], Hiroshi Mizuseki[1,2], Byungju Lee[1,\*]*

[1]Computational Science Research Center, Korea Institute of Science and Technology, 5, Hwarang-ro 14-gil, Seongbuk-gu, Seoul 02792, Republic of Korea

[2]Division of Nanoscience and Technology, KIST School, Korea University of Science and Technology, 5, Hwarang-ro 14-gil, Seongbuk-gu, Seoul 02792, Republic of Korea

[+]These authors contributed equally.

[\*]Corresponding author: Byungju Lee (blee89@kist.re.kr)

## Abstract

Recent studies have increasingly applied natural language processing (NLP) to automatically extract experimental research data from the extensive battery materials literature. Despite the complex process involved in battery manufacturing — from material synthesis to cell assembly — there has been no comprehensive study systematically organizing this information. In response, we propose a language modeling-based protocol, Text-to-Battery Recipe (T2BR), for the automatic extraction of end-to-end battery recipes, validated using a case study on batteries containing LiFePO$_4$ cathode material. We report machine learning-based paper filtering models, screening 2,174 relevant papers from the keyword-based search results, and unsupervised topic models to identify 2,876 paragraphs related to cathode synthesis and 2,958 paragraphs related to cell assembly. Then, focusing on the two topics, two deep learning-based named entity recognition models are developed to extract a total of 30 entities — including precursors, active materials, and synthesis methods— achieving F$_1$ scores of 88.18% and 94.61%. The accurate extraction of entities enables the systematic generation of 165 end-toend recipes of LiFePO$_4$ batteries. Our protocol and results offer valuable insights into specific trends, such as associations between precursor materials and synthesis methods, or combinations between different precursor materials. We anticipate that our findings will serve as a foundational knowledge base for facilitating battery-recipe information retrieval. The proposed protocol will significantly accelerate the review of battery material literature and catalyze innovations in battery design and development.

**Introduction**

In materials science, there has been a notable surge in interest towards data-driven materials informatics[1, 2]. To underpin this paradigm shift, concerted efforts have been made to obtain ample high-quality datasets. Several open databases related to materials information exist including the Materials Project[3], Open Quantum Materials Database[4], and Novel Materials Discovery[5]; however, these databases mainly consist of the results of computational studies. This insufficiency of actual experimental data can be resolved by applying natural language processing (NLP) to scientific literature[6, 7]. Research articles are meticulously curated and peerreviewed, ensuring both high quality and large quantity, from which NLP techniques can automatically extract specific information of interest[8]. In this context, text mining studies in materials science have increased in recent years, particularly in the fields of catalysts[9, 10, 11], metal-organic frameworks[12, 13], and high entropy alloys[14].

For battery materials, various NLP studies have focused on extracting information on battery materials or performance, and synthesis recipes from the literature to construct databases[15, 16, 17, 18]. Specifically, there has been a wealth of research dedicated to extracting material and property information on battery-cell assembly processes using NLP techniques such as named entity recognition (NER). For example, some pioneers suggested various literature mining protocols to extract cell-composition information such as anode, cathode, or electrolyte materials and cell-performance information such as capacity or voltage using chemistry-aware NLP techniques[15, 16, 17]. Similarly, efforts have been made to retrieve specific information such as the electrochemical characterization and cycling conditions of lithium-ion battery cells using transformer-based NER models, thereby providing a large-scale text-mined dataset with 28 entities[18]. Recently, Gou et al. suggested a document-level NLP pipeline for literature related to layered cathode materials for sodium-ion batteries[19]. The model simultaneously extracts chemical entities, electrochemical properties, and synthesis parameters.

Despite their contributions, there is still room for improvement in defining the subject of battery recipes as prior works employed limited information such as 'name of battery material' or 'synthesis recipe of battery material', which we suggest is not sufficient to represent or directly connect to battery performance data. For example, even if the same electrode material is used, differences in the cell-assembly process, e.g., using different cell types[20, 21, 22], electrode slurry recipes[20, 23, 24], separators[25, 26, 27], binders[28, 29], or electrolyte composition[30, 31, 32], greatly affect the battery performance. Therefore, to avoid ambiguity in defining battery performance, it is necessary to collect end-to-end battery recipes, where all the information from the synthesis of the electrode materials to cell assembly are gathered, before analyzing the battery performance. Notably, there has been no attempt to handle the overall process from battery materials synthesis to battery cell assembly.

In this work, we propose a language modeling-based protocol, Text-to-Battery Recipe (T2BR), for the automatic extraction of end-to-end battery material recipes from the scientific literature. As a proof of concept, we select $LiFePO_4$ cathode material, one of the most extensively studied materials in the battery field[33, 34], for our case study. First, we report machine learning (ML)-based text classification models to systematically gather papers on battery recipes. Next, we apply topic modeling such as latent Dirichlet allocation (LDA)[35], to identify paragraphs related to cathode materials synthesis and battery cell assembly. Third, NER models based on pre-trained language models are developed. The best-performing models exhibit $F_1$ scores of 88.18% and 94.61% in recognizing entities related to cathode materials synthesis and cell assembly, respectively. Our information extraction reveals trends in the usage of materials, conditions, and synthesis methods in battery experimental studies. Finally, we generate 2,840 and 2,511 sequences for two tasks based on NER results and synthesis actions[36], thereby reporting 165 end-to-end battery material recipes. Based on the recipe database, it is expected that an interactive battery recipes information retrieval system, which provides end-to-end recipes based on user inputs such as partial precursors or synthesis methods, can be developed. To the best of our knowledge, this work is the first to provide an automatic extraction of end-to-end battery material recipes from scientific literature.

# Results and discussion
## Workflow of the proposed protocol

Fig. 1 presents the comprehensive process of our T2BR protocol, which is divided into five distinct steps: (1) paper collection, (2) paper selection, (3) paragraph preparation, (4) battery recipe information extraction, and (5) battery-recipe generation. In the first step, 5,885 papers were collected by using a query consisting of several relevant keywords, such as LiFePO$_4$, on an academic search engine. Next, we developed a text-classification model to filter out irrelevant papers based on abstract information, leaving 2,174 valid papers. In the third step, we implemented topic modeling at the paragraph level, thereby identifying 2,876 and 2,958 paragraphs related to cathode material synthesis and cell assembly topics, respectively. Next, we developed NER models to extract a total of 30 entities such as the precursors, active materials, binder, atmosphere, or temperature, then revealing the usage trends using the extracted entities. Finally, we generated 2,840 and 2,511 sequences representing the process of cathode materials synthesis and cell assembly, respectively, which were used to construct 165 end-to-end battery recipes. The results for each step are described below.

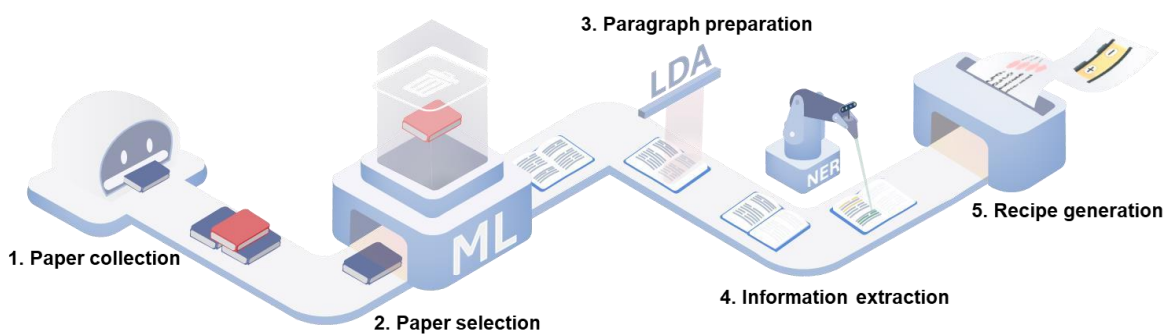

**Fig.1. Workflow of the T2BR protocol.** The following issues are considered in this workflow: (1) All the textual information of scientific literature, in addition to metadata such as paper type, publication date, or journals, is collected to filter high-quality papers. (2) Papers of interest are selected based on the abstract of papers using the ML model trained on labeled dataset. (3) Paragraph preparation is performed by an unsupervised ML model, which refers paragraph-level text information. (4) NER models are developed to extract scientific information on materials, conditions or synthesis actions, where we prepare the annotation dataset for training these models. (5) Based on the information extraction results, recipe sequences are generated and stored into our database.

**Collection and selection of battery recipe papers**
The first step of our protocol involves collecting comprehensive scientific literature on battery materials recipes. We used the ScienceDirect RESTful API, employing a search query such as ("LiFePO4" OR "lithium iron phosphate" OR "lithium ferrophosphate" OR "olivine") AND ("battery"); Our focus was on selecting documents categorized as research articles, therefore, other document types such as review articles, encyclopedias, short communications, and book chapters were excluded. This search yielded a total of 5,885 papers published up to May 2022. For each selected paper, we gathered bibliographic information, including the DOI, as well as textual information such as the title and abstract.

The results of such an information-retrieval process depend on the inclusion of specific keywords. Consequently, even if the above-mentioned keywords are mentioned in a paper, they might not necessarily pertain to battery material synthesis. To address this issue, we sampled 1,000 papers and evaluated their abstracts to determine their relevance to battery recipes. Using this dataset (true: 281, false: 719), we conducted a binary classification using Term FrequencyInverse Document Frequency (TF-IDF)-based ML models. All text classification models underwent evaluation using 5-fold cross-validation, with the optimized eXtreme Gradient Boosting (XGB) model exhibiting the highest $F_1$ score of 85.19% of five different classification models. Detailed optimization procedures for each model are provided in the Methods section. We applied the best-performing model to the remaining 4,885 papers, thereby identifying 1,893 relevant papers in addition to 281 true papers.

**Preparation of battery recipe paragraphs**
Next, we extended our analysis to the paragraphs of valid papers ($N = 2,174$). Here, the paragraphs that were too short to identify the contents (less than 200 characters) were excluded, thereby leaving 46,602 paragraphs. We applied unsupervised learning-based topic modeling, specifically LDA, to these paragraphs, to identify common topics. We adopted LDA because there is little information on how various topics are addressed, and unsupervised methods can reduce computation costs for the subsequent analysis. As a result, we were able to identify the potential distribution of topics across the full texts and to pinpoint paragraphs specifically related to battery recipes. Specifically, we identified 25 distinct topics (Fig 2.a) and interpreted the main content of each topic based on the analysis of their most frequent keywords, which revealed that two of these topics were related to battery recipes, one on the synthesis of cathode materials and the other focused on battery cell assembly (Fig 2.b–c). The topic of cathode material synthesis encompassed 2,876 paragraphs, characterized by frequent key terms such as 'solution', 'h', 'temperature', 'mixture', and 'powder'. The topic of battery cell assembly comprised 2,958 paragraphs, with frequent keywords including 'cell', 'electrode', 'electrochemical', 'cathode', 'electrolyte', and 'foil'. Thus, by employing unsupervised learning techniques such as statistical topic distribution inference, we were able to efficiently identify the main content of paragraphs related to battery recipes and accurately determine the locations of these recipe-related paragraphs within the research papers. The primary keywords for the remaining 23 topics, judged to be unrelated to battery recipes, are delineated in Table S3.

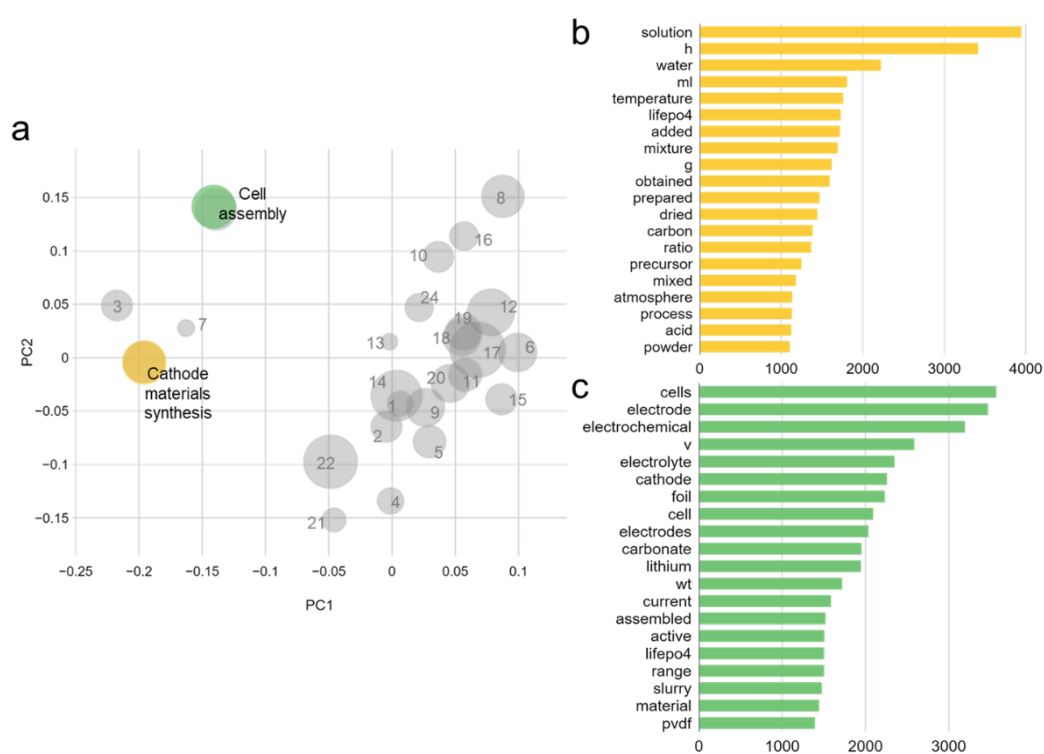

**Fig. 2. Results of filtering valid papers and paragraphs.** (a) Two-dimensional map of topics, which was obtained by applying principal component analysis to a topic-keyword distribution matrix. Here, the node size is proportional to the ratio of each topic within the entire corpus. (b–c) Frequent keywords of 'cathode materials synthesis' and 'cell assembly' topics.

**Information extraction of battery recipes**
Next, we developed two NER models to extract specific information about cathode materials synthesis and battery cell assembly. To achieve this, we created an annotated dataset where the start and end indices of each category were marked with special tags. Our deep learning-based NER models primarily consisted of bidirectional encoder representations from transformers (BERT)[37] and conditional random field (CRF)[38] layers, as illustrated in Fig. 3. For cathode materials synthesis, we identified 15 categories: precursors, temperature, target materials, time, amount, ratio, atmosphere, company, method, solvent, wash solvent, speed, solution, coating, and pH. We manually annotated 100 paragraphs, carefully reading and marking the relevant entities. For cell assembly, we defined 15 categories: amount, cathode solvent, active materials, binder, conductive agent, anode, solvent, salt, current collector, temperature, time, company, size, separator, and pressure. The descriptions and statistics of these annotations are provided in Table S2–3. We annotated 200 paragraphs to develop the NER models for this task. Specific details on the model training and their mechanisms are provided in the Methods section.

In simple terms, as illustrated in the example in Fig. 3, the first token of the input text, 'LiFePO4', is tagged as 'S-AM' for a single-word entity of the 'Active Materials' category. The NER model is trained to accurately predict the tag for each word by considering the surrounding context such as the meaning and the predicted tags of neighboring tokens. This mechanism enables the model to determine the start and end positions of words corresponding to each category, thereby facilitating the extraction of relevant information. We employed the BERT-CRF model for the NER task, utilizing various domain-specific BERT models to investigate the impact of their context-understanding abilities. The efficacy of NER is influenced by both the specific characteristics of the subject under analysis (cathode material synthesis vs. battery cell assembly) and the domain specificity of the language model's training corpus. Specifically, we tested four types of pre-trained language models: BERT[37], SciBERT[39], BatteryBERT[15], and MatBERT[40] and compared their performance in terms of $F_1$ score (Table S4–5).

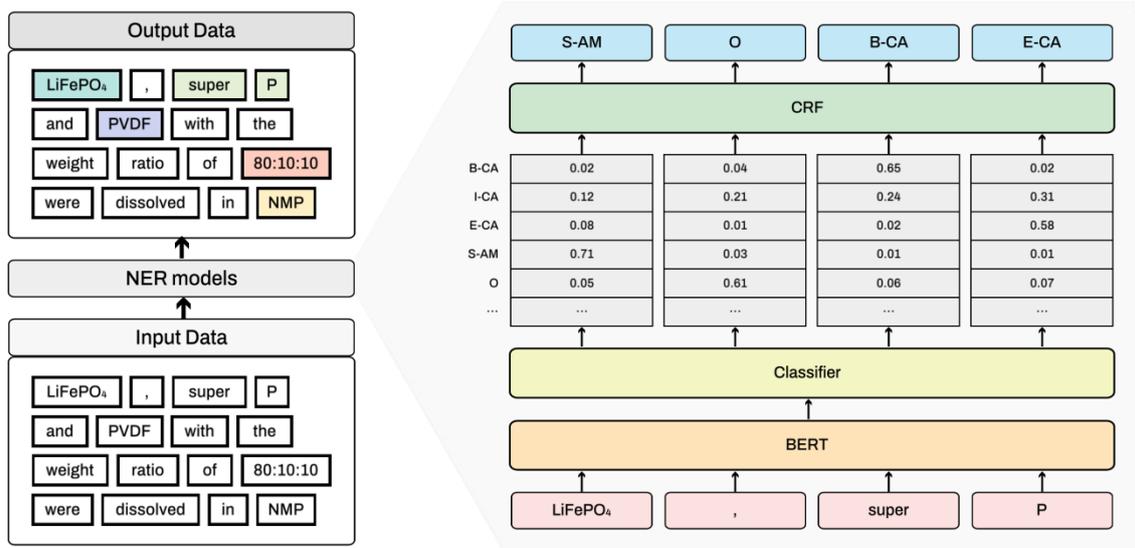

**Fig 3. Schematic illustration of our NER model.** The original text of the paper concerning battery recipes undergoes tokenization by the tokenizer, followed by the NER model, which predicts the category for each token. The NER model comprises a BERT layer for capturing the contextual meaning of each token, alongside a SoftMax function and a CRF layer designed to predict the sequence with high probability.

For the NER task focused on cathode materials synthesis, MatBERT exhibited the highest performance, achieving an average $F_1$ score of 88.18% (Fig. 4a). This superior performance can be attributed to the substantial similarities between the synthesis procedures for cathode materials and those of inorganic materials, which are well-represented in MatBERT's training corpus. Consequently, the model's tokenizer demonstrates enhanced word recognition capabilities, leading to improved NER performance in this specific context. In the category of materials information, such as precursors and target materials, MatBERT and BatteryBERT demonstrated superior performance. For instance, MatBERT achieved an $F_1$ score of 86.97% in recognizing 'target materials' entities, whereas SciBERT scored 81.97%. This superior performance of MatBERT likely stems from its specialization in materials knowledge. Conversely, for quantitative information categories such as 'temperature', 'time', and 'ratios', SciBERT and BERT show better performance. This finding suggests that domainspecific adaptation of language models may diminish their ability to recognize general numerical information.

For the cell-assembly NER task, BatteryBERT is the best-performing model, exhibiting the highest average $F_1$ score of 94.61% (Fig. 4b). We attribute the abundant battery knowledge of BatteryBERT to its superior performance, as it encompasses various terms about battery cell components, such as the anode and active materials that are exclusive to the context of battery technology. Specifically, for the anode entity, BatteryBERT achieved an $F_1$ score of 90.21%, outperforming other models such as BERT (87.22%), SciBERT (87.64%), and MatBERT (89.65%). Similarly, the BatteryBERT-based model demonstrated a superior ability to recognize 'conductive agents' entities, achieving a higher $F_1$ score (93.60%) compared to other models (BERT: 86.77%, SciBERT: 90.82%, MatBERT: 91.53%). This finding suggests that the BatteryBERT model exhibits a specialized contextual understanding of battery-related literature, enhancing its performance in identifying materials with specific roles such as 'anode' (90.21%) or 'active materials' (95.93%) within battery systems. Conversely, for categories such as salts and solvents that are relevant across numerous material domains beyond batteries, MatBERT — designed to comprehensively cover the literature on inorganic materials — demonstrated superior performance with $F_1$ scores of 94.79% and 96.74%, respectively.

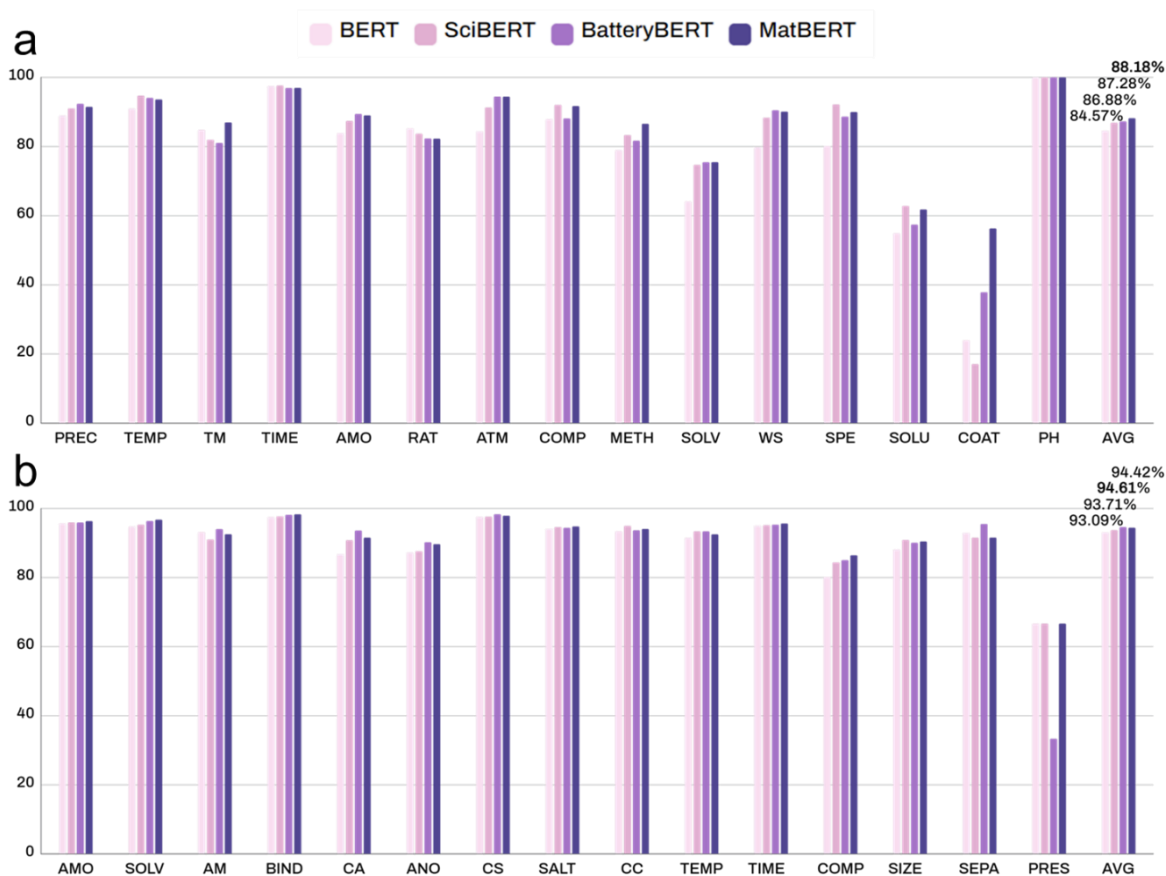

**Fig. 4. Performance of NER models for two information extraction tasks.** (a) Abbreviated categories for cathode material synthesis, i.e., precursors (PREC), temperature (TEMP), target materials (TM), atmosphere (ATM), ratio (RAT), active materials (ATM), company (COMP), method (METH), solvent (SOLV), speed (SPE), wash solvent (WS), solution (SOL) and coating (COAT). (b) Abbreviated categories for cell-assembly process, amount (AMO), solvent (SOLV), active materials (AM), binder (BIND), conductive agent (CA), anode (ANO), cathode solvent (CS), current collector (CC), temperature (TEMP), company (COMP), separator (SEPA), and pressure (PRES).

Additionally, we confirmed the potential of utilizing large language models (LLMs) for NER tasks. Detailed performance metrics and methodologies are provided in the Methods section and Supplementary Information (Table S6–7; Fig. S1). Specifically, five-shot learning with GPT-4 ('gpt-4-0416') achieved notable $F_1$ scores of 82.58% and 86.89% for information extraction tasks on cathode materials synthesis and cell assembly, respectively (Fig. S2). However, it is important to note that the results obtained from prompt engineering differ in format from those of standard NER outputs, precluding direct performance comparisons. Although the current performance of LLMs on NER tasks is somewhat limited, it can become useful when the volume of the annotated dataset is small. Furthermore, significant improvements are anticipated with advancements in prompt engineering.

**NER-based battery research trend analysis**

We applied the highest-performing NER models to the remaining paragraphs to extract all entities from battery recipe papers. Specifically, we employed the MatBERT-based NER model for 2,776 paragraphs associated with cathode material synthesis and the BatteryBERT-based model for 2,758 paragraphs related to battery cell assembly. In addition to the NER results, we extracted the synthesis action information to provide the full information of end-to-end battery recipes as sequences. To this end, we used the pre-trained text-mining toolkit for inorganic materials synthesis[36], which classifies the verbs related to synthesis action into eight categories such as 'starting', 'mixing', 'purification', 'heating', 'cooling', 'shaping', 'reaction', and 'nonaltering' based on the context. Based on the information extraction results, we were able to reveal the relationships between entities in cathode materials synthesis or battery cell assembly paragraphs (Fig. 5).

As shown in Fig. 5a, the atmosphere used for synthesizing the cathode material is predominantly Ar, followed by $N_2$, $H_2$, air, and vacuum. In summary, 77% of cathode material synthesis occurs in an Ar or $N_2$ atmosphere at temperatures between 0°C and 100°C or 600°C and 800°C, with room temperature being the most common. Above 1000°C, the synthesis is primarily conducted in an atmosphere of Ar, $N_2$, air, or $H_2$. However, less frequently, environments such as $C_2H_2$, $CH_4$, $O_2$, or vacuum are also used. Under sub-zero conditions, the synthesis primarily uses atmospheres of Ar, followed by $N_2$, inert gases, air, $H_2$, and vacuum conditions. In Fig. 5b, the combination of $LiPF_6$ with ethylene carbonate (EC) and dimethyl carbonate (DMC) predominates as the salt and solvent in most cases. EC is used due to its high dielectric constant and wide electrochemical stability window, which facilitate the dissociation of $LiPF_6$ and enhance battery stability. DMC is selected for its low viscosity and excellent electrochemical stability, which, when combined with EC, improve the electrolyte's flow properties and overall performance. In addition, EC solvent is occasionally mixed with solvents such as ethyl methyl carbonate (EMC), dimethyl ether (DME), propylene carbonate (PC), and dioxolane (DOL), whereas vinylene carbonate (VC), dimethoxymethane (DMM), dimethylformamide (DMF), and acetonitrile (CAN) are used less frequently.

In Fig. 5c, the association relationships between precursor materials and synthesis methods in battery cell assembly are visualized. From the perspective of precursors, our dataset on LFP batteries indicates that Li, Fe, and $PO_4$ sources are the most frequently extracted, with $Li_2CO_3$, $FeC_2O_4$, and $NH_4H_2PO_4$ being the most commonly used. Most studies adopted the solid-state method for synthesizing uniformly formed LFP particles, primarily using $Li_2CO_3$, $FeC_2O_4$, or $NH_4H_2PO_4$ as precursor materials. For hydrothermal methods, LiOH, $FeSO_4$, or $H_3PO_4$ precursors are used, whereas $H_3PO_4$ and LiOH are frequently selected in the solvothermal method as well. They are selected because of their ability to act as a versatile reactant under elevated temperatures and pressures in aqueous or solvent environments, facilitating controlled crystallization and the formation of desired nanostructures or complex compounds with tailored properties. The sol-gel method was mainly employed for handling citric acid or $NH_4$, whereas the precipitation, rheological phase, or polymerization method was sometimes used for $FeSO_4$, $NH_4$, and S, respectively. In Fig. 5d, the dependency relationships of precursor materials in battery recipes are analyzed. In summary, $Li_2CO_3$ and $NH_4$ are frequently used together, because of their ability to efficiently provide lithium ions and facilitate the formation of homogeneous and high-purity cathode materials. In addition, there are dominant combinations such as LiOH − $H_3PO_4$, $FeSO_4$ − LiOH, $FeC_2O_4$ − $NH_4$, $FeSO_4$ − $H_3PO_4$, and $Li_2CO_3$ − $FeC_2O_4$. In addition to these results, we analyzed the relationships between temperature − time and temperature − action in cathode materials synthesis and binder − conductive agent and temperature − action in battery cell assembly (Fig. S3).

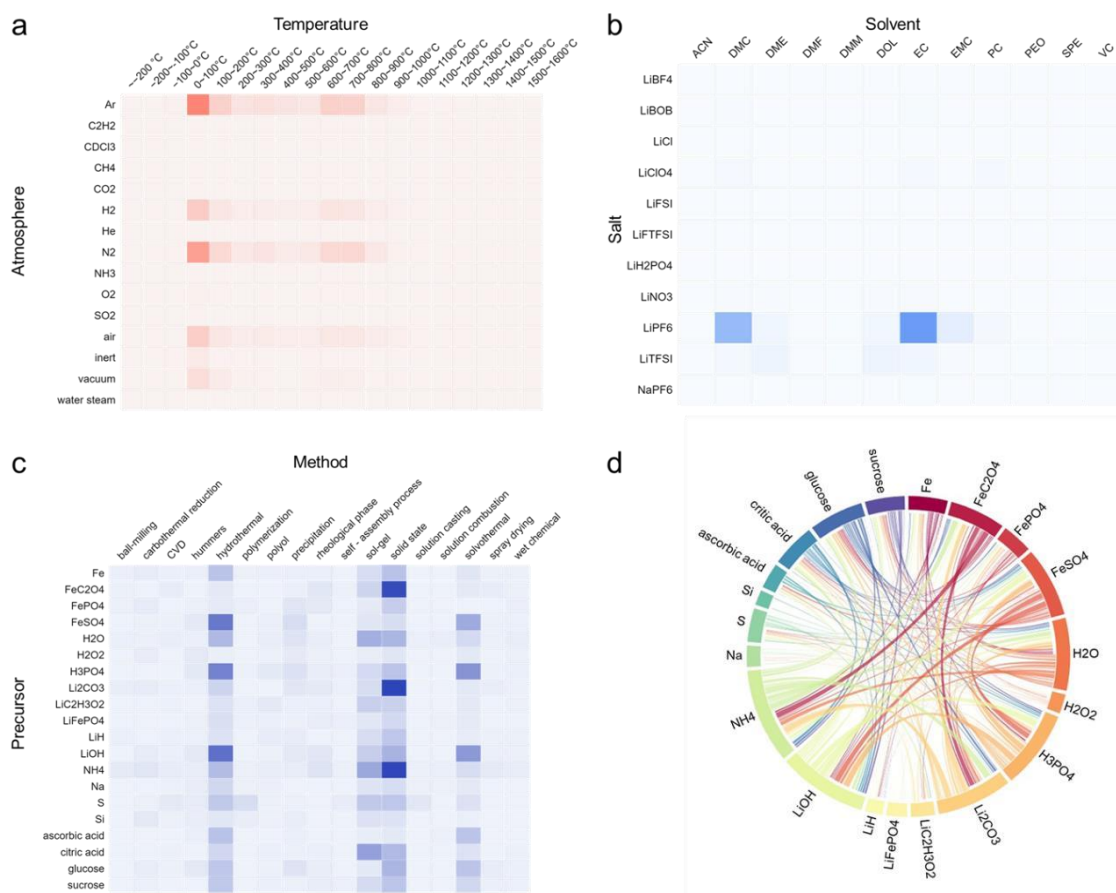

**Fig. 5. Outcomes of NER-based research trend analysis.** In heatmaps, the color represents the normalized number of relevant papers; (a) Relationships between atmosphere and temperature in cathode materials synthesis. (b) Relationships between salt and solvent in cell assembly. (c) Relationships between synthesis method and precursor materials in cathode materials synthesis. Here, CVD refers to chemical vapor deposition. (d) Dependency relationships between precursor materials.

**Battery recipe pattern analysis and retrieval**

Here, we assumed that the sequential mention of synthesis actions in the text represents the order of the synthesis process. The synthesis actions and NER results are displayed according to the order of sentence appearances in the sequence data. Finally, we obtained 2,840 sequences from paragraphs detailing the cathode materials synthesis and 2,511 sequences from paragraphs describing battery cell assembly. Next, we aimed to uncover potential causal relationships between synthesis actions by probabilistically analyzing the previously derived cathode material synthesis sequences ($N = 2,840$) and cell assembly sequences ($N = 2,511$). As a result of analyzing sequences of synthesis actions in cathode material synthesis paragraphs, the sequence with the highest probability is identified as < 'starting' → 'mixing' → 'purification' → 'heating' > (Fig. 6a). The reason for this high probability is that the synthesis of cathode materials typically begins with the preparation of raw materials ('starting'), followed by their combination to ensure uniformity ('mixing'). Subsequent purification steps are crucial to remove impurities that could affect material performance, and finally, heating is applied to induce the necessary chemical reactions and phase transformations. An analysis of 2,511 sequences of synthesis actions in cell-assembly process paragraphs identified the most probable sequence as < 'starting' → 'mixing' → 'non-altering' → 'purification' > (Fig. 6b). The reason for this high probability is that the cell-assembly process typically begins with the preparation of initial components ('starting'), followed by their combination to ensure homogeneity ('mixing'). The non-altering step involves procedures that do not change the chemical nature of the components, such as coating slurry to current-collector layers. Finally, purification steps are essential to remove any contaminants that could compromise the performance and longevity of the cell.

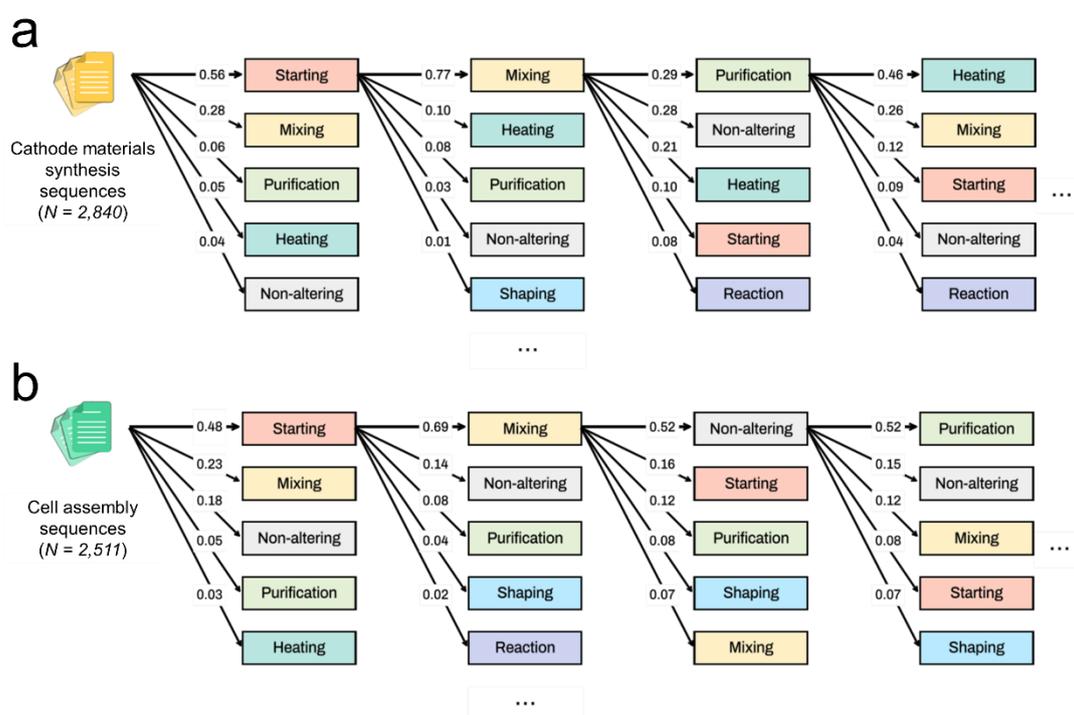

**Fig. 6. Results of sequence probability modeling of battery recipe sequences.** The sequence probabilities were calculated by cumulatively determining the conditional probabilities of each step

based on the sequences found in the cathode material synthesis paragraphs or cell assembly process paragraphs. (a–b) 2,840 and 2,511 sequences, related to cathode materials synthesis and cell assembly paragraphs respectively, are analyzed.

Next, we tried to identify end-to-end battery recipes, which encompass the entire process from material synthesis to cell assembly, by linking and filtering the two-type recipes. For this task, the following post-processing steps were conducted. First, we verified whether the source papers of the material synthesis recipe and the cell assembly recipe were the same. Next, we confirmed whether the target material resulting from the cathode material synthesis sequence and the active material, which is the starting material for the cell assembly sequence, were the same. Finally, we ensured that the precursor and synthesis methods were clearly specified in the given recipes, thereby identifying 165 end-to-end recipes. The reason why the number of end-to-end recipes is relatively small is because not all LFP battery studies cover the entire process from material synthesis to cell assembly. In the collected dataset, numerous instances were found where only the cathode synthesis process was detailed, primarily concentrating on material synthesis and characterization. For instance, when the research objective involves analyzing the morphological characteristics of specific materials such as $FePO_4$ and $LiFePO_4$, the aim is to understand the structure, size, and thermal behavior of these materials[41, 42]. Consequently, the focus is on their physical and chemical properties, with no evaluation of the electrochemical performance of the battery cell. Furthermore, several studies concentrated exclusively on the synthesis process of $LiFePO_4$ particles and their properties during synthesis[43, 44]. Conversely, in instances where only the cell assembly process was described, the cathode was often procured commercially, with only the source being specified. These studies typically omitted descriptions of the cathode synthesis process[45, 46, 47, 48].

Based on this recipe database, an interactive battery recipe information retrieval system can be developed, as illustrated in Fig. 7. If precursor materials are limited and only solid-state synthesis methods are available, users can search our database to find relevant recipes, including cathode synthesis, cell assembly or end-to-end types. Searching with a query such as "(('sucrose'). PREC.) AND (('solid state'). METHOD) AND (('end-to-end'). TYPE)" provides the following end-to-end recipe: In Step 1, the target material, $LiFePO_4/C$, is synthesized from raw materials such as $LiH_2PO_4$, $FeC_2O_4·2H_2O$, 5% sucrose, and 5% citric acid. In Step 2, a slurry is prepared using LiFePO$_4$ ('active material'), Super P ('conductive agent'), and PVDF ('binder'), which is then coated onto aluminum foil to form the cathode. Next, the anode is prepared using lithium foil, and a microporous PE film is inserted between the two electrodes to serve as the separator. Finally, the electrolyte, consisting of LiPF$_6$ mixed with EC and DEC solvents, is added to complete the battery cell assembly. In this way, by using certain precursor elements or synthesis method conditions as input, it is possible to provide the complete recipe for material synthesis or cell assembly.

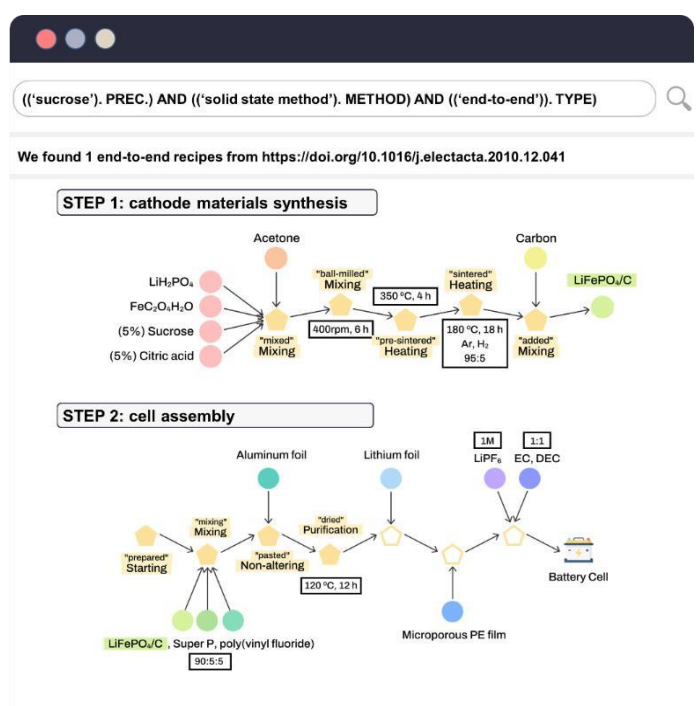

**Fig. 7. Implementation of an interactive battery recipe information retrieval system.** Our recipe information retrieval system allows materials scientists to search for battery recipes by specifying chosen precursors or synthesis methods. When integrated into a web service, the system provides visualization capabilities for the recipes.

## Conclusion

In this work, we aimed to systematically and automatically extract end-to-end battery recipes from the scientific literature using a language modeling-based protocol, i.e., T2BR. First, we developed ML-based text classification models to discern the papers related to battery recipes from the information retrieval results, leading to the filtering of 2,174 valid documents with a high $F_1$ score of 85.19%. Next, we conducted topic modeling at the paragraph level, efficiently identifying 2,876 and 2,958 paragraphs about cathode materials synthesis and cell-assembly processes, respectively. We developed two deep-learning-based NER models, each designed to extract 15-type entities— one model focused on cathode materials synthesis (e.g., precursors, target materials) and the other targeting entities associated with cell-assembly processes (e.g., active materials, anode). These models exhibited high average $F_1$ scores of 88.18% and 94.61%, respectively, enabling the automatic extraction of battery recipe entities from the remaining approximately 5,500 paragraphs. In addition, we extracted the synthesis action using a materials-aware NLP toolkit[36], thereby generating 165 sequences representing the overall process of battery recipes. To the best of our knowledge, this study is the first to collectively extract end-to-end battery recipes from large-scale scientific literature, paving the way for a generalized approach to the knowledge base construction of battery materials.

We acknowledge several limitations of the current study and propose directions for future research. First, our analysis is based on a limited dataset of $LiFePO_4$ battery literature, collected exclusively from a single search engine. Consequently, some reports based on information extraction results may exhibit bias. However, our protocol is adaptable and can be applied to an expanded dataset that includes other battery systems[49, 50] such as lithium-ion batteries consisting of $LiCoO_2$ and $LiMn_2O_4$ cathode materials, as demonstrated by several examples in Fig. S4–5. The second limitation arises from the lack of connection between battery recipes and the electrochemical performance of the batteries. Our protocol enables the extraction of battery recipe information, not providing quantitative information on the electrochemical profile of batteries such as the voltage–capacity curve, charge/discharge curve, cycle life, energy density, and current–voltage curve. Considering that the long-term goal is to identify the optimal battery recipe by linking our end-to-end battery recipes with performance data, it is essential to analyze additional information from tables and figures[51] as well as extract relationships between entities from text. Finally, we have identified potential areas for performance enhancement in the proposed protocol, likely attributable to discrepancies between the pre-defined categories and actual data. The diverse and complex terminology employed by materials scientists when referring to battery components may contribute to this performance degradation. Consequently, it is imperative to use an annotation dataset that ensures both quality and diversity. For categories with limited datasets (e.g., method, solvent, solution, coating, pressure), augmenting existing data, rather than merely increasing tagging sets, may serve as a viable alternative. Moreover, exploring other state-of-the-art models, such as pointer networks for NER, could potentially improve contextual understanding and final output performance. For instance, applying fine-tuning of LLMs presents an opportunity to develop an NER model specific to battery recipe extraction[52, 53]. Considering the presence of exceptional cases, such as the intertwining of cell assembly and material synthesis information in a single paragraph, generative models could potentially offer a solution in the future.

## Methods

**Paper-level text classification models**
We classified battery-recipe-related papers from the information retrieval results in a systematic manner. Initially, we manually reviewed the abstract and title information of 1,000 randomly selected papers to determine their relevance to battery recipes. This process resulted in 281 relevant and 719 irrelevant papers. We used this labeled dataset to develop a paper classification model. First, we applied TF-IDF to represent the text as vectors using the scikitlearn, generating a 1,000 × 10,592 matrix. We then developed five classification models using AutoML in the H2O module[54]: random forest (RF), logistic regression (LR), gradient boosting machine (GBM), multi-layer perceptron (MLP), and XGB. To optimize the hyperparameters for each ML model, we conducted a grid search based on 5-fold cross-validation, using the $F_1$ score as the performance evaluation metric.

The optimal parameters identified were as follows: the number of trees and maximum depth were 50 and 20 for the RF. The family was set as binomial distribution, and the minimum lambda, beta epsilon, theta, and stopping tolerance were set as 0.0001, 0.0001, 1e-10, and 0.001, respectively, for LR. The following parameters were set for GBM: sample rate = 0.8, learning rate = 0.1, stopping tolerance = 0.001, stopping metric = log loss, maximum depth = 15, and number of trees = 50. Three hidden layers of 100 nodes, an Adam optimizer with a 0.005, a rectified linear unit as the activation function, and a dropout rate of 0.5 for MLP were used. For XGB, the following parameters were used: number of trees = 110, maximum depth of the trees

= five levels, learning rate = 0.03, a scale of positive weight = 2, minimum child weight = 2, and minimum split loss = 3. Among these models, the optimized XGB model demonstrated the highest performance (Table S8). We then applied this model to the remaining 4,885 papers. Combining the 281 manually labeled papers with the prediction results, where 1,893 papers were classified as relevant, we identified a total of 2,174 papers related to battery recipes.

**Paragraph-level topic modeling**

We conducted paragraph-level topic modeling using Python libraries such as Natural Language Toolkit (NLTK)[55] and gensim[56]. First, NLTK was used for pre-processing such as tokenization and stopwords elimination, where common articles such as 'a', and 'the', and pronouns such as 'this' and 'that' were excluded. Next, genism was employed to develop the LDA model. LDA is a probabilistic model that provides insights into the topics present within a given document[35]. It estimates topic-specific word distributions and document-specific topic distributions from datasets consisting of documents and their constituent words. This inference process relies on the assumption of distributions following the Dirichlet distribution, a common practice in Bayesian models of multivariate probability variables. In essence, LDA posits that words within a document are generated based on the joint distribution of topic word distributions and document topic distributions and utilizes Gibbs sampling to infer these distributions from the observed word distribution within the document.

LDA involves two hyperparameters: $\alpha$ and $\beta$. The former determines the density of document-topic relationships, whereas the latter indicates the density of topic-word relationships. Their higher values lead to more uniform probabilities across topic distributions, whereas lower values emphasize specific topic distributions. We set these parameters as 5.0 and 0.01. In addition, we determined the number of topics as 25 based on coherence and perplexity scores. Perplexity gauges the efficacy of a probability model in predicting observed values, with lower values indicating superior document–model alignment. Coherence, on the other hand, evaluates the semantic consistency within topics[57]. As modeling accuracy increases, topics tend to aggregate semantically related terms. Consequently, by assessing the similarity among primary terms, we ascertain the semantic coherence of topics. Through exhaustive testing across topic numbers ranging from 1 to 40, we identified 25 topics characterized by optimal balance between low perplexity and high coherence. For the visualization of topic modeling results, we used the LDAvis Python library, which provides an interactive web-based visualization.

**Pre-trained language model configuration**

For the NER tasks, we employed the BERT-CRF model. We tested a range of pre-trained language models, including BERT ('bert-base-uncased'), SciBERT ('scibert_scivocab_uncased'), MatBERT ('matbert-base-uncased'), and BatteryBERT('batterybert-uncased'). These models were trained on distinct domain corpora, resulting in variations in their word recognition and contextual understanding capabilities. Specifically, BERT was trained on general knowledge sourced from books and Wikipedia text (~3300M words), whereas SciBERT was trained on research papers from the fields of biology, medicine, and computer science (~3170M words). MatBERT was trained on materials science research papers (~8.8B words), and BatteryBERT was fine-tuned based on BERT specifically using papers from the field of battery materials (~1870M words). BERT-based models are trained with two tasks such as masked language modeling, which predicts masked words within a given sequence, and next sentence prediction, which discerns relationships between sentences. Their contextual understanding capabilities vary depending on the corpus used for training, which directly affects their performance across NLP tasks in different domains. Furthermore, each BERT model utilizes a distinct tokenizer, as they rely on the byte pair encoding algorithm. Consequently, the level at which consecutive character sequences, appearing with a certain frequency in the corpus, are recognized as a single token varies depending on the corpus used. As depicted in Fig. S6, the word 'LiFePO4' can be tokenized either as a single entity or segmented into multiple tokens ('LiFe', 'PO', '4').

**NER model development**

After tokenization, we annotated the tokens using the IOBES tagging scheme, which classifies each entity into subtypes to indicate whether a token is inside (I), outside (O), at the beginning (B), or the end (E) of multi-token entities as well as single-token entities (S). This scheme is effective for handling compound words, as it provides additional information about the boundaries of named entities. This tagging scheme can introduce certain constraints in sequence labeling tasks. For example, an I-tag cannot appear at the beginning of a sentence, and an OI pattern is invalid. In a B-I-I pattern, the named entity must remain consistent; for instance, B-AM can be followed by I-AM or E-AM, but not by I-CA. To address these sequence labeling challenges, we employed a CRF layer as the final layer of the NER models. A CRF is a type of SoftMax regression that transforms categorical sequential data into a format suitable for SoftMax regression, subsequently used to predict sequence vectors.

After model configuration, the dataset with annotations is split into training, validation, and test sets with a ratio of 8:1:1 with stratified sampling. In training NER models, we conducted an exhaustive grid search to optimize the hyperparameters, which are determined as follows: the maximum number of epochs of 50, applying early stopping with the patience of 10, the learning rate of 1e-3, batch size of 5, and optimizer of RangerLars, which is the composite of RAdam, LARS, and Lookahead. We conducted 10-fold shuffle split crossvalidation, which is effective in adjusting imbalanced datasets, and verified the performance of the NER models. Next, when evaluating the NER performance for each category, we adopted a lenient evaluation criterion by applying boundary relaxation, considering the diversity and complexity of entity expressions in the materials science and battery recipe domains. Specifically, if any component word of a compound term belonging to a specific category was correctly identified, it was deemed a correct match.

**Post-processing of information extraction results**

After extracting specific information from the scientific literature, we normalized the entities written in natural language in a qualitative but systematic manner. First, based on frequent entities for each category, we constructed a dictionary of chemical substances. Next, we identified the entities with similar meanings but different entities by conducting pairwise comparisons of frequent entities. Consequently, entities representing the same active materials such as $LiFePO_4/C$, and $LiFePO_4/Carbon$ were normalized into the more frequently appearing one such as $LiFePO_4/C$, whereas binder-type entities such as polyvinylidene fluoride and PVDF were unified into PVDF. Finally, these normalization results were used to analyze the battery research trends in depth. When normalizing synthesis time information, all the values were converted to seconds and transformed using a logarithmic scale, divided into 10 intervals for analysis. 'Overnight' was estimated as 8 h, and 'few' and 'several' were approximated as

5. For normalizing synthesis temperature information, Kelvin temperatures were converted to Celsius. When temperature ranges were provided, we checked if they fell within 100-degree intervals. For example, if the extracted entity was '150–220', both the 0–100 and 100–200 intervals were marked.

## Data and code availability
The source codes and related datasets are available at https://github.com/KIST-CSRC/Text-to-BatteryRecipe or can be obtained from a corresponding author upon request.

## Author contributions
B.L. conceived the idea. B.L. and H.M. supervised the project. D.L. contributed to the preparation of annotation dataset, model development, and visualization. J.C. contributed to data collection, methodology design, model development/validation, result analysis, data/code repository management, and visualization. J.C. and B.L. wrote and edited the manuscript, and all authors contributed to the discussion and writing of the manuscript.

## Conflicts of interest
There are no conflicts to declare.

## Acknowledgments
This work was supported by the National Research Foundation of Korea funded by the Ministry of Science and ICT (NRF-2021M3A7C2089739) and Institutional Projects at the Korea Institute of Science and Technology (2E33211 and 2Z07160). We would like to express our sincere gratitude to Nayeon Kim, a Ph.D. student at the Korea Institute of Science and Technology, for her valuable assistance in the editing of our figures.